\documentclass[preprint,12pt]{elsarticle}

\usepackage{graphicx}      
\usepackage{amsmath,amssymb,amsfonts}
\usepackage{algorithmic}
\usepackage{textcomp}
\usepackage[utf8]{inputenc}
\usepackage{color}

\usepackage{lipsum}

\usepackage{amsbsy}
\usepackage{multirow}

\usepackage{subfig}
\usepackage{amsfonts, amsmath, amsthm, amsbsy,amssymb,bm,mathtools} 
\usepackage{graphicx} 		
\usepackage{transparent}	
\newsubfloat{figure}		
\graphicspath{{figures/}} 	

\usepackage[dvipsnames]{xcolor}

\usepackage{lineno,hyperref}
\usepackage{mathtools}
\usepackage{verbatim}

\allowdisplaybreaks[1]
\journal{Neurocomputing}

\begin{document}
\newcommand{\eric}[1]{#1}

\newcommand{\ericc}[1]{\textcolor{ProcessBlue}{--EA: #1}}
\begin{frontmatter}



\title{Investigation of Proper Orthogonal Decomposition for Echo State Networks}


\author[UFSC]{Jean~Panaioti~Jordanou}
\author[UFSC]{
        Eric~Aislan~Antonelo}
\author[UFSC]{Eduardo~Camponogara}
\author[Texas]{Eduardo Gildin}


\address[UFSC]{Department of Automation and Systems, Federal University of Santa Catarina, Florianopolis, 88040-900, Santa Catarina, Brazil}
\address[Texas]{Harold Vance Department of Petroleum Engineering, Texas A\&M University, College Station, 77843-3116, Texas, United States of America}         

\begin{abstract}
Echo State Networks (ESN) are a type of Recurrent Neural Network that yields promising results in representing time series and nonlinear dynamic systems.
Although they are equipped with a very efficient training procedure, Reservoir Computing strategies,  such as the ESN, require high-order networks, i.e., many neurons, resulting in a large number of states that are magnitudes higher than the number of model inputs and outputs.
A large number of states not only makes the time-step computation more costly but also may pose robustness issues, especially when applying ESNs to problems such as Model Predictive Control (MPC) and other optimal control problems. 
One way to circumvent this complexity issue is through Model Order Reduction strategies such as the Proper Orthogonal Decomposition (POD) and its variants (POD-DEIM), whereby we find an equivalent lower order representation to an already trained high dimension ESN.
To this end, this work aims to investigate and analyze the performance of POD methods in Echo State Networks, evaluating their effectiveness through the Memory Capacity (MC) of the POD-reduced network compared to the original (full-order) ESN.
We also perform experiments on two numerical case studies: a NARMA10 difference equation and an oil platform containing two wells and one riser.
The results show that there is little loss of performance comparing the original ESN to a POD-reduced counterpart and that the performance of a POD-reduced ESN tends to be superior to a normal ESN of the same size. Also, the POD-reduced network achieves speedups of around $80\%$ compared to the original ESN.
\end{abstract}


\end{frontmatter}



\section{Introduction}

Recurrent Neural Networks (RNN) are very relevant in applications related to modeling real-world phenomena when time-dependent data are available \cite{nl_sys_ident}, \cite{Bishop}, and are considered universal approximators of dynamic systems.
As RNNs are nonlinear, their training suffers from issues such as local minima, slow training, and the so-called ``fading gradient'' problem \cite{deep_goodfellow}, which is a numerical problem inherent in Backpropagation Through Time (BPTT) \cite{Mozer}, the algorithm used to calculate an RNN gradient.
While some solutions focus on solving the fading gradient problem by changing the RNN structure, such as the Long Short-Term Memory (LSTM) network \cite{lstm}, or the gated recurrent unit \cite{deep_goodfellow}, another flavor of RNN is worthy of attention: Reservoir Computing (RC).
RC simplifies the learning by dividing the RNN into two parts: a high-dimensional recurrent nonlinear layer (the reservoir) with fixed, randomly generated weights and an adaptive readout output layer, which computes an instantaneous linear combination of the dynamic reservoir states \cite{Jaeger2007335}.
The output-layer weights are trained through linear least squares, overcoming the problems related to nonlinear training and BPTT.
\textit{Reservoir Computing} became a unifying term for the frameworks of Liquid State Machines \cite{Maass2011} and Echo State Networks (ESN) \cite{Jaeger2007335}, both of which are methods for RNN training independently developed.  

ESNs follow the general reasoning of Reservoir Computing: they adopt an architecture with a dynamic reservoir with fixed weights that projects the input to a high-dimensional space and a trainable static readout output layer.
The dynamic reservoir needs to have many neurons \cite{Jaeger2007335} and the so-called Echo State property, which refers to the stability properties of the network.
There are many successful applications of ESNs, such as: learning complex goal-directed robot behaviors \cite{Antonelo2014}, fuel cell lifetime prediction \cite{Mezzi2021}, wind speed prediction \cite{Bai2021}, medium voltage insulators classification \cite{Stefenon2022}, forecasting power system load using an ensemble deep ESN \cite{Ruobin2022}, power systems prediction with enhanced ESN that employ logistic mapping and bias dropout for reservoir weights generation
\cite{Wang2022neucom}, and prediction of the daily maximum temperature in the Melbourne airport with multi-reservoir ESN and an encoding and decoding scheme \cite{Li2022}.
The large number of dynamic states in the reservoir is an essential characteristic, as the output, being a linear combination of them, can represent a more extensive repertoire of dynamics.
However, using ESNs as dynamic models for problems such as optimization and MPC (Model Predictive Control) \cite{camacho} may be an issue since the higher the number of states in the ESN is, the larger the optimization problem.
Because the number of states in the ESN heavily dominates the number of inputs and outputs in such applications, a large reservoir size renders the optimization problem inherently larger and harder to solve. 

As ESNs are high-dimensional, model order reduction methods can find equivalent ESN models with a considerably smaller number of states but which still keep the properties and performance of the original high-dimensional ESN. 
To that end, we count on Proper Orthogonal Decomposition (POD) \cite{Chatu2010}, which applies Singular Value Decomposition (SVD) to find an optimal linear transformation that represents the state space of a large dynamical system in a more compact form.
POD is already widely used to reduce the number of states of large dynamical models, especially phenomenological models such as a gas reservoir simulator \cite{Wang2018} with tens of thousands of variables.
However, POD has one disadvantage concerning nonlinear systems: although the method can reduce the number of states, it does not reduce the computation number of nonlinear functions.
There are developments of interpolation methods, such as the Discrete Empirical Interpolation Method (DEIM) \cite{Chatu2010}, to mitigate the issue by pivoting and approximating the nonlinear portion of the given model computation.
Both POD and DEIM can find lower-dimensional networks that are equivalent to the original ESN and, thus,
have the potential to alleviate the computational burden of simulations that depend on the size of the trained ESN.

The main objective of this work is to experiment with the use of POD and DEIM to obtain a reduced-order equivalent for an already-trained ESN.
For such end, we apply the reduction given by POD in three different contexts: 
a Memory Capacity (MC) \cite{memory_jaeger} evaluation experiment;
a NARMA10 difference equation \cite{Sakemi2020}; and a simulated oil platform containing two gas-lifted oil wells and one riser \cite{jordanou_eaai}.
Additionally, we have shown results using DEIM-based reduction for the ESN in the first and last experiments mentioned above.
We compare the performance of the reduced ESN to the original (non-reduced) ESN in the three experiments and another ESN with the same size as the reduced ESN in the MC and NARMA experiments.

In this context, our main contributions are two-fold: (1) we have developed efficient computational frameworks for implementing large echo-state network models in a variety of applications, which is achieved via model-order reduction (MOR) techniques; and (2) 
we have assessed the trade-offs between low-complexity reservoir models, resulting from the application of model order reduction (MOR), and the large baseline model in terms of numerical accuracy. The low-complexity models, despite their relatively small state-space dimensions, demonstrate comparable representation power to the large baseline model.
As such, our work contributes to this nascent field of applications of MOR strategies to reservoir computing, which can potentially improve computational performance in modeling, control, and optimization. 
    Specifically, the findings of our work are the following:
\begin{itemize}
    \item The memory capacity of an ESN reduced by POD is generally higher than that of a non-reduced ESN of equivalent size. This difference in memory capacity is more significant as the desired ESN gets smaller in size.


\item Given two echo state networks with the same number of states, the ESN obtained from POD reduction is likelier to perform better in a given task. This property is more evident and relevant when the desired reservoir is small.

\item By employing a MOR method on ESNs, this work shows that small ESNs are robust and performant, improving their suitability for real-time or embedded applications with memory limitations.

\item DEIM reduction alone for ESNs does not achieve satisfactory results compared to pure POD reductions. 

\end{itemize}

In broader terms, the main implication of these findings is that a smaller version of an ESN, obtained by model order reduction,  can achieve nearly equivalent behavior to the original (and larger) ESN,  thus making dynamic reservoirs more compact.
%
%
  The new model can serve as a proxy model in optimization and predictive control, as an observer, and in other related tasks, addressing the issue of computational cost in a reservoir consisting of a large number of internal states (reservoir size), which can be orders of magnitude larger than the number of inputs and outputs.
\color{black}

This paper is organized as follows: Section \ref{sec:rel_work} contains related works, Section \ref{sec:esn} presents the Echo State Networks, Section \ref{sec:mor} describes POD and DEIM, Section \ref{sec:app} reports on the case studies and experimental testing for the reduced ESN, and Section \ref{sec:conclusion} concludes the work.


\section{Related Work} \label{sec:rel_work} 

%
In the following, we will discuss works in the literature that address the issue of reducing the model size in reservoir computing.
One of them is \cite{Sakemi2020}, where they propose reducing the number of states by considering the output as a linear combination of the states at different instants in time, comparing to an original ESN through the Information Processing Capacity (IPC) metric, and also applying the proposal to a NARMA system and the generalized H\'enon-map.
The solution raises the effective number of states as a multiple of the delay or ``drift-state'' number utilized.
   
The architecture is very hardware-friendly, easing the computation compared to a standard ESN. 
  Another example is the work \cite{Whiteaker2022}, where they propose to employ the controllability matrix of the ESN as a means to find a so-called minimal ESN, which would be the ESN with the smallest reservoir that could reproduce the task at hand.
They train the ESN for a particular task, obtain the controllability matrix at given points, and define its rank as a new candidate reservoir size.
An extensive search procedure is then performed to find the optimal ESN at that size; however, there is no direct connection between the larger and the smaller ESN.
In summary, the method in \cite{Whiteaker2022} proposes a useful way of finding a minimal reservoir for a task.
In comparison, our work follows a different direction: reducing the \eric{size of the network} through POD.
Another work \cite{Liu2022} proposes a different approach to reducing reservoir size, which calculates the correlation between each neuron and eliminates the reservoir neurons with the highest correlation.

The necessary large number of reservoir states in an ESN implies a complex computational model, therefore works such as \cite{Yang2022} employ methods of so-called ``network size reduction,'' which perform multi-objective optimization on the output weights and minimize not only the least-square error but also the number of non-zero elements in the output weights.
Enforcing sparseness is ideal for simplifying computations with the ESN.
Another work that follows this line of reasoning is \cite{deterministic_esn}, where they enforce a minimum complexity ESN by forcing the ESN reservoir to follow a deterministic form (i.e., a circular reservoir).

In \cite{Lkse2017}, they propose to add the reservoir dimensionality reduction into the architecture via Principal Component Analysis (PCA) and calculate the output layer based on the PCA output instead of the reservoir states. 
They affirm that this enhances the dynamic properties of the resulting ESN concerning the system identified and improves the network generalization capabilities.
Also, applying dimensionality reduction in the states renders the ESN a tool for dynamic system analysis.
In this sense, our POD-ESN method is similar to PCA regarding obtaining the new state space but goes beyond \cite{Lkse2017} by embedding the reduction achieved in the reservoir's state update equation. In other words, the reservoir recurrent simulation is executed in the reduced state space with POD-ESN, which does not happen in \cite{Lkse2017}.

Another approach of reduction in reservoir computing, not involving POD, is proposed in \cite{Halus2020}.
Their idea involves procedurally removing neurons according to the output weight value, which they curiously discovered that the network performance improves (given the Lorentz system as an application) by removing the neurons associated with large output weights.
They thoroughly analyze the effect of removing different types of nodes in the ESN.
%

\section{Echo State Networks (ESN)} \label{sec:esn}
An ESN is a type of recurrent neural network with useful characteristics for system identification \cite{Jaeger2007335}, 
as it represents nonlinear dynamics well and the training consists in solving a linear least-squares problem of relatively low computational cost when compared to nonlinear optimization.

\subsection{Model}

\begin{figure}[!t]  
	\centering
	\includegraphics[width=0.50\textwidth]{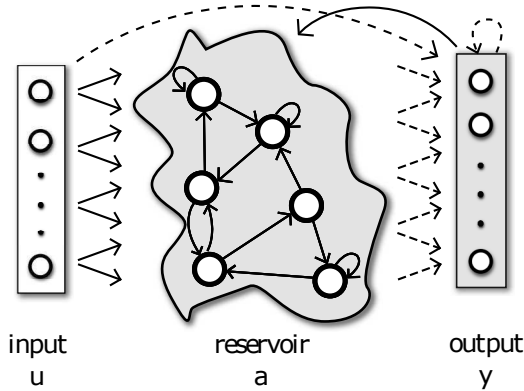}
	\caption{Representation of an Echo State Network, one of the possible models in Reservoir Computing.
		Dashed connections (from Reservoir to Output Layer) are trainable, while solid connections are fixed and randomly initialized. 
  This figure was obtained from \cite{Antonelo2017}.
	}
	\label{fig:echostate}
\end{figure}

Proposed in \cite{Jaeger2004,Jaeger2001a}, the ESN is governed by the following discrete-time dynamic equations:
\begin{align}
\mathbf{x}[k+1] &= (1-\gamma)\mathbf{x}[k] \nonumber \\ &~ + \gamma \mathbf{f}(\mathbf{W_r^r}\mathbf{x}[k] + \mathbf{W_i^r}\mathbf{u}[k] + \mathbf{W_b^r} + \mathbf{W_o^r}\mathbf{y}[k]) \label{eq:esn}\\
\mathbf{y}[k+1] &= \mathbf{W_r^o}\mathbf{x}[k+1] \label{eq:output_esn},
\end{align}
where: the state of the reservoir neurons at time $k$ is given by $\mathbf{x}[k]$;
the current values of the input and output neurons are represented by $\mathbf{u}[k]$ and $\mathbf{y}[k]$, respectively;
$\gamma$ is called leak rate \cite{Jaeger2007335}, which governs the percentage of the current state $\mathbf{x}[k]$ that is transferred into the next state $\mathbf{x}[k+1]$.
The weights are represented in the notation $\mathbf{W_{from}^{to}}$, with ``$\mathbf{b}$'', ``$\mathbf{o}$'', ``$\mathbf{r}$'', and ``$\mathbf{i}$'' meaning the bias, output, reservoir, and input neurons, respectively; 
and  $f = \tanh(\cdot)$ is an activation function widely used in the literature, also called a base function in system identification theory \cite{nl_sys_ident}.
Fig. \ref{fig:echostate} depicts a standard architecture of an echo state network. 

The network has $N$ neurons in the reservoir, which is the dimension of $\mathbf{x}[k]$ and is typically orders of magnitude higher than the number of network inputs.
As long the network training uses regularization, $N$ can be as large as needed, but at the expense of increased computation time to update the reservoir states as defined in \eqref{eq:esn}.
According to \cite{memory_jaeger}, the ESN with no output feedback connections (the output does not affect the state), which is given by $\mathbf{W_o^r}$, has a memory capacity ($MC$) bounded by the number of neurons in the reservoir ($MC \leq N$), assuming the use of linear output units.

The recurrent reservoir should possess the so-called Echo State Property (ESP)  \cite{Jaeger2001a}, i.e., a fading memory of its previous inputs, meaning that influences from past inputs on the reservoir states vanish with time.
The ESP is guaranteed for reservoirs with $\tanh(\cdot)$ as the activation function, provided that the singular values of $\mathbf{W_r^r} < 1$. 
However, this condition limits the richness of the reservoir's dynamical qualities, which discourages its use in practice. 
Note that all connections going to the reservoir are randomly initialized, usually according to the following steps:
\begin{enumerate}
	\item Every network weight is initialized from a normal distribution $\mathcal{N}(0,1)$.
	
	\item $\mathbf{W_{r}^{r}}$ is scaled so that its spectral radius  $\rho$ (Eigenvalue with the largest module) characterizes a regime able to create reservoirs with rich dynamical capabilities.
	Setting $\rho < 1$ in practice often generates reservoirs with the ESP \cite{Jaeger2007335}. 
	However, reservoirs with $\rho > 1$ can still have the ESP since the effective spectral radius may still be lower than 1 \cite{Ozturk,Verstraeten2009}.

	\item $\mathbf{W_i^r}$ and $\mathbf{W_b^r}$ are multiplied by scaling factors $f_i^r$ and $f_b^r$, respectively, affecting  the magnitude of the input.
\end{enumerate}

These scaling parameters, $\rho$, $f_i^r$, and $f_b^r$ are crucial in the learning performance of the network, having an impact on the nonlinear representation and memory capacity of the reservoir
\cite{memorynl}.
Also, low leak rates allow for higher memory capacity in reservoirs, while high leak rates favor quickly varying inputs and outputs.
The settings of these parameters should be such that the generalization performance of the network (loss on a validation set) is enhanced. 

\subsection{Training}
While in standard RNNs all weights are trained iteratively using gradient descent \cite{Mozer}, ESNs restrict the training to the output layer $\mathbf{W_r^o}$.
Because the echo state property does not emerge with output feedback $\mathbf{W_o^r}\mathbf{y}[k]$, this work favors reservoirs without feedback from the output, i.e., $\mathbf{W_o^r}=0$.
Also, the inputs do not interfere directly with the output, as systems with direct transmission are less smooth and more noise-sensitive.
To train an ESN, the input data $\mathbf{u}[k]$ is arranged in a matrix $\mathbf{U}$ and the desired output $\mathbf{d}[k]$ in vector $\mathbf{D}$ over a simulation time, where each row $\mathbf{u}^T$ of $\mathbf{U}$ corresponds to a sample time $k$ and its columns are related to the input units. 
For the sake of simplicity, we assume that there are multiple inputs and only one output.
The rows of $\mathbf{U}$ are input into the network reservoir according to each sample time, thereby creating a state matrix $\mathbf{X}$ containing the resulting state sequence.
Then, we apply the Ridge Regression algorithm \cite{Bishop} by using $\mathbf{X}$ as the input data matrix and $\mathbf{D}$ as the output data matrix or, in this case, a vector as we assumed single output. 
Ridge Regression results in solving the following linear system:
\begin{equation}
(\mathbf{X}^T\mathbf{X} - \lambda \mathbf{I})\mathbf{W_r^o} = \mathbf{X}^T\mathbf{D}
\label{eq:training},
\end{equation}
where $\lambda$is the Tikhonov regularization parameter, which penalizes the weight magnitude and avoids overfitting.
There are also methods to apply least-squares training online \cite{nl_sys_ident}, but this work does not use these algorithms.

\section{Model Order Reduction} \label{sec:mor}

In this section, \eric{we propose Model Order Reduction (MOR) methods for reducing the reservoir dimensionality in ESNs, specifically the Proper Orthogonal Decomposition (POD) and the Discrete Empirical Interpolation Method (DEIM).}
We also propose a strategy for correcting the steady-state error introduced in ESNs by MOR methods.

\subsection{Proper Orthogonal Decomposition}

The Proper Orthogonal Decomposition is a method to find a linear transformation \cite{Chen} $\mathbf{T}$ for a given system that maps a high-dimensional state space  into a reduced one, namely:
\begin{equation}
    \mathbf{x} = \mathbf{T}\mathbf{z}
\end{equation}
where $\mathbf{x}$ is a vector of dimension $n$ and $\mathbf{z}$ is a vector of dimension $m \ll n$, so that $\mathbf{T}\in\mathbb{R}^{n\times m}$.

The transformation itself is akin to a similarity transformation, with the main difference being that $\mathbf{T}$ lacks an inverse for not being a square matrix.
However, the $\mathbf{T}$ resulting from POD is orthonormal ($\mathbf{T}^T\mathbf{T} = \mathbf{I}$), so the transpose is used in place of an inverse.

To find $\mathbf{T}$, we gather snapshots of the states in a given dynamical system response, akin to gathering data in a machine learning problem.
The columns of the snapshot matrix $\mathbf{X}\in\mathbb{R}^{n\times N}$ are the states $\mathbf{x}[k]\in \mathbb{R}^{n}$, where $N$ is the number of snapshots such that $N\geq n$.
Then, we wish to minimize the error induced by projecting the original state onto the reduced space and back, which leads to the following error function: 
\begin{equation}
    E(\mathbf{T}) = \sum_{k = 1}^{N}\bigl(\mathbf{x}[k] - \mathbf{T}\underbrace{\mathbf{T}^T\mathbf{x}[k]}_{\mathbf{z}[k]}\bigr)^2
\end{equation}
The second term is $\mathbf{x}$ \textit{projected} onto the reduced space of $\mathbf{z}$, and then \textit{lifted} back.
The optimal $\mathbf{T}$ is obtained through singular value decomposition (SVD) \cite{Sun2017}, decomposing $\mathbf{X}$ in the following form:
\begin{equation}
    \mathbf{U_{svd}}\mathbf{\Sigma}\mathbf{V}^T = \mathbf{X}
\end{equation}
where $\mathbf{U_{svd}}$ contains the left singular vectors and has dimension $n \times n$, $\mathbf{\Sigma}$ contains the singular values and has dimension $n \times N$, with only $n$ non-zero columns.
We consider that $\mathbf{\Sigma}$ is sorted from the largest to the smallest singular value.
POD does not use the right singular vector matrix $\mathbf{V}$.

The transformation $\mathbf{T}$ that minimizes $E(\mathbf{T})$ is found by concatenating the columns with the $m$ largest corresponding singular values from $\mathbf{U_{svd}}$.
We seek a truncation so that the reduced system energy is close to the original, measured by:
\begin{align} \label{eqn:energy_contribution}
    \epsilon &= \sum\limits_{j=1}^{m}\epsilon_j
    & \epsilon_j &= 
    \sigma_j/{\sum\limits_{i=1}^n\sigma_i}
\end{align}
where $\epsilon$ is the total energy contribution of the singular values maintained in the reduced-order model, $\sigma_j$ is the $j^{th}$ highest singular value, $\epsilon_j$ is the energy contribution of that given singular value, and $m$ is the reduced state dimension.
The energy contribution of the remaining singular values in the reduction is a metric on how close the reduced-order model is to the original system regarding information.
For this work, we measure the energy contribution of each singular value of the original signal and truncate $\mathbf{U_{svd}}$ to obtain $\mathbf{T}$ so that $\epsilon$ reaches a desired energy contribution value (\textit{e.g.}, $\epsilon = 0.95$, so that the reduced system has $95\%$ of the original system's energy).
In other words, the reduced-order model carries $\epsilon$ information of the original system.
After obtaining $\mathbf{T}$ for the dimension reduction through the process above, the reduced ESN dynamics can be expressed as follows:
  \begin{subequations}
  \label{eq:reduced_esn_pod}
\begin{align}
\mathbf{z}[k+1] &= (1-\gamma)\mathbf{z}[k] \nonumber \\ &~~ + \gamma \mathbf{T}^{T}\mathbf{f}(\mathbf{W_r^r}\mathbf{T}\mathbf{z}[k] + \mathbf{W_i^r}\mathbf{u}[k] + \mathbf{W_b^r}) \label{eq:reduced_esn_state}\\
\mathbf{y}[k+1] &= \mathbf{W_r^o}\mathbf{T}\mathbf{z}[k+1] \label{eq:output_reduced_esn},
\end{align}
\end{subequations}

We can observe from the operation $\mathbf{T}^T\mathbf{f}(\cdot)$ that the reduced-order ESN does not reduce the number of computations by only performing POD on it.
In fact, to compute the element-wise $\tanh$, $\mathbf{T}$ brings the dimension
back to the original state space size, which is to be reduced again with $\mathbf{T}^{T}$, increasing the number of computations.
This computational increase is inherent in POD for nonlinear systems and will be dealt with by the method described in the next section.


\subsection{Discrete Empirical Interpolation}

The Discrete Empirical Interpolation Method (DEIM) is an approximation method to circumvent the POD computation issue \cite{Chatu2010}, which consists of state projection and lifting operations to compute state transitions in the reduced-order model.
The core idea of DEIM is to approximate the nonlinear term of a dynamic system as a polynomial interpolation that resembles the strategy employed in POD.
Given the following discrete-time nonlinear system:
\begin{equation}  
    \mathbf{x}[k+1] = \mathbf{A}\mathbf{x} +  \mathbf{f}(\mathbf{x}[k]), 
\end{equation}
where the nonlinear function is elementwise, meaning  that 
\begin{equation}
\mathbf{f} = \bigl (f(\mathbf{x}), f(\mathbf{x}), \dots, f(\mathbf{x})\bigr)  
\end{equation}
for a given function $f$ such as ${\tt tanh}$. Notice that the system is divided into linear and nonlinear portions.
Applying the POD ($\textbf{x} = \textbf{T}\textbf{z}$) into such a system yields:
\begin{equation} \label{eqn:deim_sys}
    \mathbf{z}[k+1] = \mathbf{T}^T\mathbf{A}\mathbf{T}\mathbf{z}[k] + \mathbf{T}^T\mathbf{f}(\mathbf{T}\mathbf{z}[k])
\end{equation}

The nonlinear mapping $\mathbf{f}$ of the dynamic system can be approximated as follows: 
\begin{equation} \label{eq:DEIM:eq01}
    \mathbf{P}^T\mathbf{f}(\mathbf{T}\mathbf{z}[k]) \approx \mathbf{P}^T\mathbf{U}\mathbf{c}[k]
\end{equation}
where $\mathbf{U} \in \mathbb{R}^{n\times m}$, which is obtained from the same POD as $\mathbf{T}$, however with a different number $m$ of singular vectors, with $n$ being the number of states, and $\mathbf{P}$ is a pivoting matrix of the same dimension as $\mathbf{U}$. 
DEIM interprets that a linear combination, with basis $\mathbf{U}$ and the elements $\mathbf{c}[k]$ as function coefficients, approximates the elementwise function $\mathbf{f}$.

After obtaining $\mathbf{U}$ from $\mathbf{U_{svd}}$, we then obtain $\mathbf{P}$ with the following procedure \cite{Chatu2010}:
\begin{enumerate}
    \item The index and value of the largest element of the first left-singular vector is stored in a list. $\mathbf{P}$ starts as a column matrix with the only non-zero element being the value $1$ at the row corresponding to this index.  
    
    \item For each column $l \geq 2$ of the POD left-singular vectors (where $\mathbf{\widetilde{U}}_{l}$ is a matrix with the first $l-1$ columns of $\mathbf{U}$):
    \begin{enumerate}
        \item find $\mathbf{c}$ where $(\mathbf{P}^T\mathbf{\widetilde{U}}_{l})\mathbf{c} = \mathbf{P}^T\mathbf{u}_l$, where $\mathbf{u}_l$ is the left-singular vector corresponding to the $l^{th}$ column of $\mathbf{U}$. 
    
        \item Calculate $\mathbf{r} = \mathbf{u}_l - \mathbf{\widetilde{U}}_{l}\mathbf{c}$ and store the maximum absolute value and index of $\mathbf{r}$ in a list. Add a new column to $\mathbf{P}$ according to the obtained index.
        
    \end{enumerate}
    
    \item Output: Pivoting matrix $\mathbf{\mathbf{P}}$ according to the order dictated by the index list obtained.
\end{enumerate}

This procedure guarantees that $\mathbf{P}^T\mathbf{\widetilde{U}}_{l}$ is always nonsingular; thus $\mathbf{c}$ is the unique solution to the linear system in step 2 \cite{Chatu2010}. Letting $\mathbf{U}$ be the matrix of left singular values obtained from the procedure, it follows from \eqref{eq:DEIM:eq01} that:
\begin{equation} \label{eqn:interpolation_deim}
    \mathbf{c}[k] = (\mathbf{P}^T\mathbf{U})^{-1}\mathbf{P}^T\mathbf{f}(\mathbf{T}\mathbf{z}[k])
\end{equation}

The result from \eqref{eqn:interpolation_deim} leads to the DEIM function interpolation:
\begin{equation} \label{eqn:deim_approx}
    \hat{\mathbf{f}}(\mathbf{T}\mathbf{z}[k]) \approx \mathbf{U}(\mathbf{P}^T\mathbf{U})^{-1}\mathbf{P}^T\mathbf{f}(\mathbf{T}\mathbf{z}[k])
\end{equation}

This function approximation has an ${\ell}_2$ error bound of the following form \cite{Chatu2010}:
\begin{equation}
     e_{{\ell}_2}(\mathbf{f}) \leq \|(\mathbf{P}^T\mathbf{U})\|_2\|(\mathbf{I} - \mathbf{U}\mathbf{U}^T)\mathbf{f}(\mathbf{T}\mathbf{z}[k])\|
\end{equation}
where, in turn:
\begin{equation}
    \|(\mathbf{P}^T\mathbf{U})\|_2 \leq (1 + \sqrt{2n})^{m-1}\|\mathbf{u}_1\|_\infty^{-1}
\end{equation}
with $\mathbf{u}_1$ being the first column of $\mathbf{U}$ and 
 $n$ being the number of original states.

The main advantage of DEIM is that, as $\mathbf{f}$ is an element-wise nonlinear function, the following equality holds:
\begin{equation}\label{eqn:deim_approximation}
    \underbrace{\mathbf{U}(\mathbf{P}^T\mathbf{U})^{-1}\mathbf{P}^T}_{\mathbf{T}_1\in \mathbb{R}^{n\times n}}\underbrace{\mathbf{f}(\mathbf{T}\mathbf{z}[k])}_{\mathbf{f}:\mathbb{R}^n\rightarrow \mathbb{R}^n} =
      \underbrace{\mathbf{U}(\mathbf{P}^T\mathbf{U})^{-1}}_{\mathbf{T}_2\in \mathbb{R}^{n\times m}}\underbrace{\mathbf{f}(\mathbf{P}^T\mathbf{T}\mathbf{z}[k])}_{\mathbf{f}:\mathbb{R}^m\rightarrow \mathbb{R}^m}  
\end{equation}
 The difference between the right-hand side and left-hand side of this equation is better seen in a compact form,
\begin{equation*}
    \mathbf{T}_1\mathbf{f}(\mathbf{T}\mathbf{z}[k]) = \mathbf{T}_2\mathbf{f}(\mathbf{P}^T\mathbf{T}\mathbf{z}[k])
\end{equation*}
where $\mathbf{T_1}$ has $n$ columns, which yields the same computation problem as the original Galerkin projection, whereas $\mathbf{T_2}$ has  $m$ columns, which is the reduced state space.
This simple difference grants huge computational savings since the online calculations would be performed in terms of the reduced dimension $m$, $m\ll n$, which mitigates the computation issues regarding the POD method.

The DEIM-approximated reduced order ESN has the form obtained by applying DEIM from Eq. \eqref{eqn:deim_approximation} into the already reduced ESN at \eqref{eq:reduced_esn_pod}:
\begin{subequations}
\begin{multline}
\mathbf{z}[k+1] = (1-\gamma)\mathbf{z}[k] \\ 
  + \gamma \mathbf{T}^T\mathbf{T_2}\mathbf{f}\left (\mathbf{P}^T\mathbf{W_r^r}\mathbf{T}\mathbf{z}[k] + \mathbf{P}^T\mathbf{W_i^r}\mathbf{u}[k] + \mathbf{P}^T\mathbf{W_b^r}\right ) \label{eq:reduced_esn}
\end{multline}
\begin{equation}
\mathbf{y}[k+1] = \mathbf{W_r^o}\mathbf{T}\mathbf{z}[k+1] \label{eq:output_reduced_esn_deim},
\end{equation}
\end{subequations}
The property $\mathbf{P}^T\mathbf{f}(\cdot) = \mathbf{f}(\mathbf{P}^T)$ holds for elementwise operations, which justify the matrix placement in the DEIM reduced-order ESN.

\subsection{Stability Loss in DEIM} \label{sec:DEIM-stability}

According to \cite{Selga2012}, a contractive linear system is guaranteed to retain stability when applying POD for model order reduction; therefore, if the ESN is contractive, the POD-ESN is guaranteed to retain stability.
However, DEIM has no such property.
Assume an equilibrium point $\mathbf{x_{eq}}$ of the ESN, and a fixed input $\mathbf{u}$,
\begin{equation}
\mathbf{x_{eq}} = \mathbf{f}(\mathbf{W_r^r}\mathbf{x_{eq}} + \mathbf{W_i^r}\mathbf{u}+ \mathbf{W_b^r})
\end{equation}
 and its reduced mapping $\mathbf{z_{eq}}=\mathbf{T}^T\mathbf{x_{eq}}$. The Jacobian of the full and reduced order model are:
\begin{align}
J(\mathbf{x_{eq}}) &= (1-\gamma)\mathbf{I} + \gamma \mathbf{f}'(\mathbf{g}(\mathbf{x_{eq}}))\mathbf{W_r^r}\\
J(\mathbf{z_{eq}}) &= (1-\gamma)\mathbf{I} + \gamma \mathbf{T}^T\mathbf{f}'(\mathbf{g}(\mathbf{T}\mathbf{z_{eq}}))\mathbf{W_r^r}\mathbf{T}
\end{align}
where:
\begin{equation}
 \mathbf{g}(\mathbf{x}) = \mathbf{W_r^r}\mathbf{x} + \mathbf{W_i^r}\mathbf{u} + \mathbf{W_b^r}
\end{equation}

Since $\mathbf{f'}$ is a diagonal matrix where each element belongs to the interval $(0,1]$ for being the elementwise derivative of the $\tanh$ function, the stability of the ESN in both cases is governed by $\mathbf{W_r^r}$ at an equilibrium point. 
Also, as per \cite{Selga2012},  the POD reduction retains the stability of the ESN. Summing up, the original and reduced-order ESNs are stable provided that the spectral radius of $\mathbf{W_r^r}$ is smaller than $1$.

With DEIM, however, the stability is not retained, as shown by calculating the Jacobian of an ESN reduced by both POD and DEIM:
\begin{multline}
\mathbf{J_{DEIM}}(\mathbf{z}) = (1-\gamma)\mathbf{I} + \gamma \mathbf{T}^T\mathbf{U}(\mathbf{P}^T\mathbf{U})^{-1}\mathbf{f}'(\mathbf{P}^T\mathbf{g}(\mathbf{T}\mathbf{z}))\mathbf{P}^T\mathbf{W_r^r}\mathbf{T}
\end{multline}

Notice that the term $(\mathbf{P}^T\mathbf{U})^{-1}$ can amplify the Jacobian to the point that the ESN dynamic system has an unstable eigenvalue, despite POD-ESN being stable.
This term represents the pivoting of the truncated singular vectors associated with DEIM.


\section{Applications} \label{sec:app}

This section presents results from experiments with reduced-order ESNs for three case studies, along with a preliminary analysis on the singular values of the ESN snapshots.

\subsection{Preliminary Study: Energy contribution distribution in Echo State Networks}

\begin{figure}
    \centering
    \includegraphics[width=0.85\textwidth]{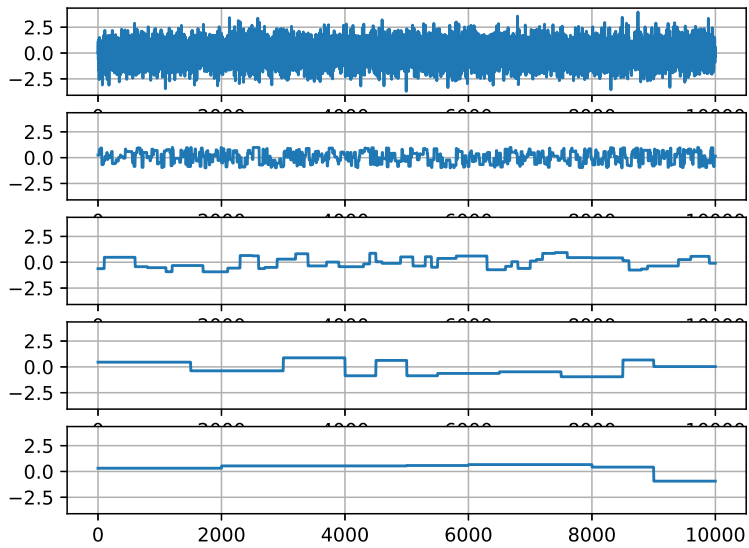}
    \caption{One-dimensional input signals for the reservoir energy contribution distribution experiment.
    White noise (top), APRBS signals (usually used in identification tasks): with a minimum period of 10, 100, 500, and 1,000 timesteps, respectively, from second topmost plot to bottom.
    }
    \label{fig:svd_input}
\end{figure}

POD and DEIM originate from applying SVD into the ESN state response matrix, obtained from exciting the ESN's reservoir with an input signal.
Thus, the SVD does not depend on the output layer.
To test the influence of input signals into the singular values of the state snapshots, we initialize $20$ different single-input ESN reservoirs and apply SVD into the snapshots of the response obtained from the reservoir, given as inputs with $10,000$ timesteps:

\begin{itemize}
    \item A white noise following the normal distribution $\mathcal{N}(0,1)$.
    \item Four different APRBS (Amplitude-modulated Pseudo-Random Binary Signal) random stair signals, defined by their minimum period, i.e., $10$ timesteps, $100$ timesteps, $500$ timesteps, and $1,000$ timesteps.
    \item A concatenation in time of all the signals above.
\end{itemize}
The input signals for the experiments are shown in Figure \ref{fig:svd_input}.
Note that this discussion concerns only the state dynamics of the reservoir; therefore, it is neither dependent on the identified system nor on the output weights.

After exciting the ESN with the signals mentioned above, one at a time, we perform SVD of the resulting ESN state response snapshots and plot the energy contribution $\epsilon_j$ associated with each singular value, sorted from highest to lowest according to Eq. \eqref{eqn:energy_contribution}.
All the reservoirs employed for this experiment are fully leaked ($\gamma = 1$), have $500$ neurons, a spectral radius $\rho = 0.99$, and a value $0.1$ for both input scaling and bias scaling.

\begin{figure}
    \centering
    \includegraphics[width=0.85\textwidth]{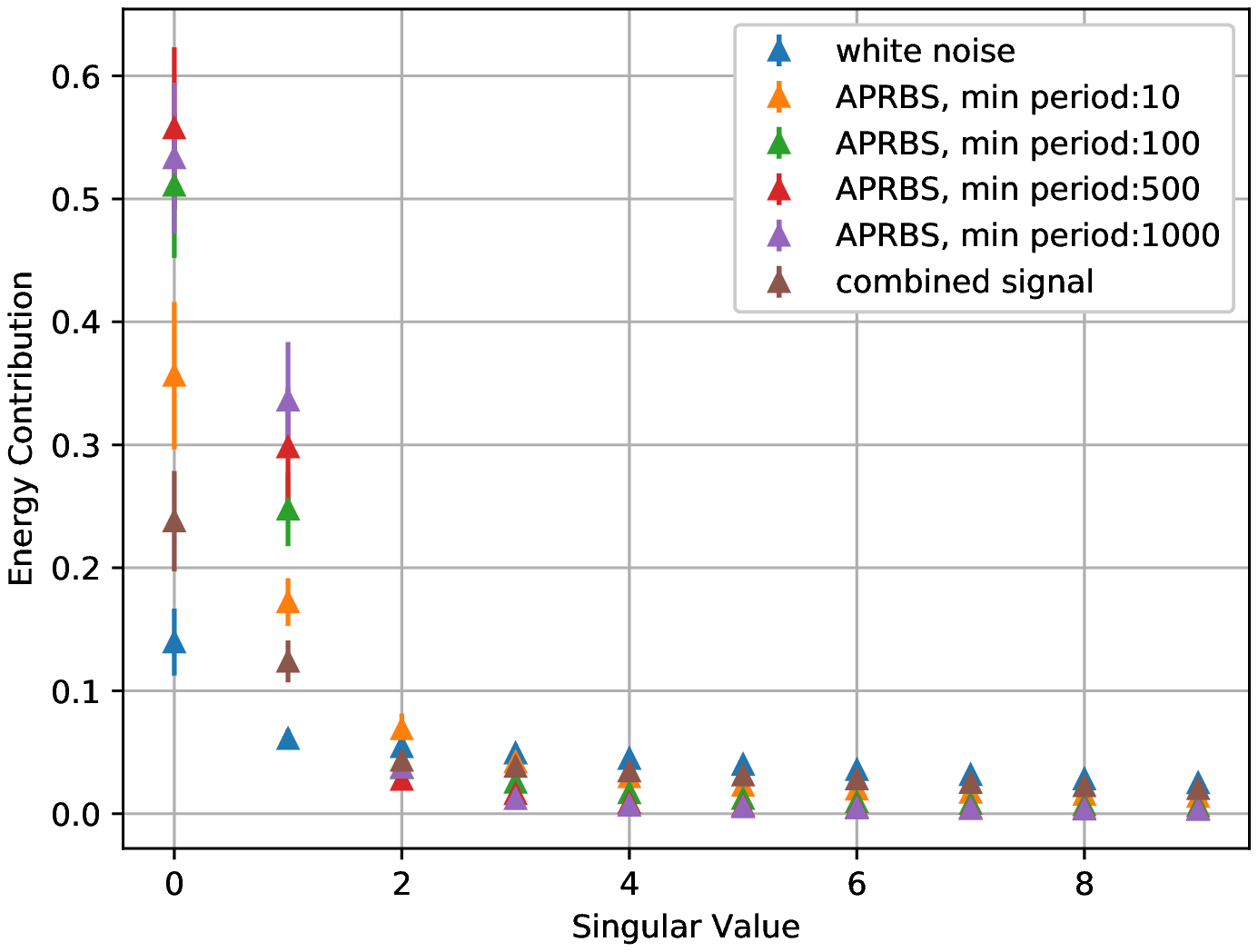}
    \caption{Mean and Standard deviation of the first ordered $10$ singular values (with $0$ corresponding to the highest and $9$ to the lowest)  obtained from the snapshots of $20$ different ESN reservoirs. Each color corresponds to a different input signal fed to the ESN reservoir, shown in Fig. \ref{fig:svd_input}.}
    \label{fig:ec_profile}
\end{figure}

Figure \ref{fig:ec_profile} showcases the mean and standard \eric{deviation of the} energy contribution of the $10$ highest singular values for each state snapshot
considering $20$ randomly initialized reservoirs.
We infer from this result that the singular values become more evenly distributed the higher the frequencies of the input signal are.
As the white noise is a signal with heavy high-frequency information, we expect the ESN state response to having a more even energy contribution distribution among the singular values. 

Meanwhile, the lower frequency signals have the energy contribution concentrated about the highest magnitude singular value.
In fact, real-life dynamic systems work as low pass filters \cite{Chen} and, therefore, they are expected to have lower frequency information. 
The slower the system dynamics are, the larger the minimum period of an APRBS signal needs to be, which directly affects the singular value profile of the model order reduction.

This experiment implies that, since the distribution of the energy contribution depends entirely on the input signal frequency, the number of states pruned by MOR is higher for cases with low-frequency dynamics.
After all, since the energy contribution is more concentrated on the first singular values, the number of columns pruned is higher than when the singular values are more evenly distributed (as in the case of high-frequency signals like white noise).
As an easy example, the highest energy contribution singular value for the APRBS signal with a minimum period of $1,000$ timesteps contributes more to the total energy of the snapshots than the sum of the $10$ highest singular values for the white noise shown in the plot.

\subsection{Memory Capacity Evaluation}

Short-term Memory Capacity (MC) is a well-known metric for Echo State Networks \cite{memory_jaeger} \eric{that} measures how well an ESN can remember past inputs and general dynamic storage capacity.
MC serves as a performance measurement for ESN reservoirs which is obtained from the following procedure:
\begin{itemize}
    \item For an arbitrary $n$, train a single-input, single-output Echo State Network so that the input is a given white noise $\mathbf{\eta}[k]$, and the output is the same white noise delayed $n$ timesteps $\mathbf{\eta}[k-n]$.
    In layman's terms, the ESN is supposed to ``memorize'' the input \eric{from} $n$ timesteps ago.
    
    \item Obtain the correlation coefficient $R_n$ for the training with an arbitrary $n$,
    \begin{equation}
        R_n = \frac{{\tt cov}(y_{esn},\mathbf{\eta}[k-n])}{{\tt var}(\mathbf{y_{esn}}){\tt var}(\eta[k-n])}
    \end{equation}
    where ${\tt cov}(\cdot)$ is the covariance operator, $y_{esn}$ is the single ESN output,  ${\tt var}(\cdot)$ is the variance operator, and, therefore, $R_n$ is merely the determination coefficient for a given delay $n$.
    \item The memory capacity is calculated, in theory, as:
    \begin{equation}
        MC = \sum_{n=1}^{\infty}R_n
    \end{equation}
\end{itemize}

The MC of an ESN was mathematically proven to have an upper bound in its number of neurons $N$ \cite{memory_jaeger}, which means that it is directly related to the number of network neurons.

For this work, we propose an experiment to compare the memory capacity of the reduced-order model of the ESN, and the original ESN, since the number of neurons is the upper bound for MC.
Because it is impossible to run infinite training experiments, we define the memory capacity for this experiment as follows:
\begin{equation}
        MC = \sum_{n=1}^{N_{MC}}R_n
    \end{equation}
where $N_{MC}=100$ is a sufficiently large number to measure the memory capacity of the network. As preliminary tests show, after a given $n$, the determination coefficient converges to a low value. Therefore, the information regarding memory capacity is more concentrated in the lower $n$ spectrum, endorsing the limited number of experiments ($N_{MC} = 100$) for comparison purposes.

\subsubsection{POD Reduction}

We ran the memory capacity experiment for different numbers of neurons ($N = \{400,600,800, 1000, 1200, 1400, 1600, 1800, 2000, 2200 \}$) with an Energy Cutoff (EC) of $1\%$, $5\%,$ and $10\%$.
After initializing the ESN reservoir at random, we perform model order reduction for $12$ different reservoirs in each configuration.
We then measure the mean and standard deviation for the memory capacity of these twelve runs while also obtaining the range of the reduced dimension for a given energy cutoff. 
This analysis allows us to measure the memory capacity drop for the model order reduction and assess how reservoir-dependent the order-reduction procedure is.

All reservoirs analyzed are fully leaked ($\gamma = 1.0$) and have input and bias scaling at $0.1$.
Also, the reservoir spectral radius is $\rho = 0.99$.

\begin{figure}
    \centering
    \includegraphics[width=0.78\textwidth]{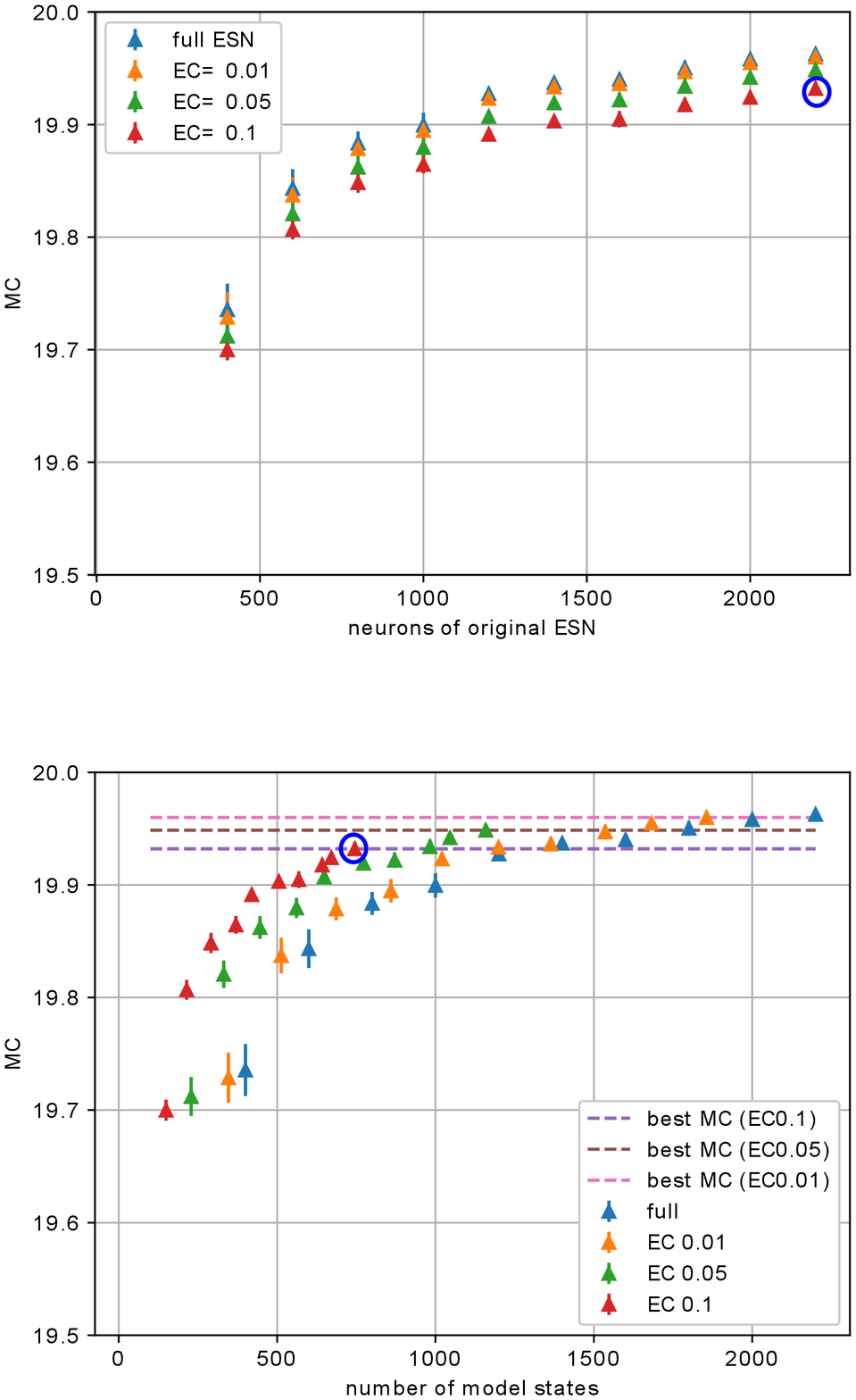}
    \caption{Plot of the memory capacity as a function of the number of neurons of the original network (upper plot), and as a function of the number of states (lower plot). 
    Each point is colored according to the energy cutoff of the POD-ESN that obtained the MC shown (points in blue are the MC obtained from full ESNs). EC means the energy cutoff of the applied POD.}
    \label{fig:mc_experiment_podvsfull}
\end{figure}

Figure \ref{fig:mc_experiment_podvsfull} showcases the results of the Memory Capacity experiments when
performing MOR at the tested ESNs given different energy cutoffs, depicting both mean and standard deviation of the $12$ runs.

The first plot depicts the number of ESN neurons before applying POD to a given network.
It shows the expected drop in MC resulting from applying MOR with more energy cutoff.

Meanwhile, the second plot portrays the MC as a function of a given network's exact number of states after performing MOR through POD.
As MC progresses monotonically, given the number of states, either in an ESN or in a given MOR of that ESN, it becomes easy to map a point of the second plot into the first one: for example,  the last red point 
\eric{(from left to right)} 
of both plots (marked within a blue circle)
have the same memory capacity since they
correspond to the same network/EC configuration.
Thus, the MOR of an ESN with $2200$ neurons (first plot) has roughly $750$ states (second plot) at 1\% energy cutoff.

As per the previous section, since this experiment traditionally employs a white noise signal, the drop in the number of reduced states is not very significant; however, the drop in MC is still small, given that a large number of states were still cut off (even in the case of $10\%$ energy cutoff for the $2200$ neuron network, the number of states was reduced to almost a third).
In fact, the second plot shows that a POD-reduced network ends up being more powerful in terms of MC than a full (non-reduced) ESN with the same number of states: when we compare an ESN with a given reservoir size to a POD-reduced network from a larger ESN with the same number of states as that ESN reservoir size, the POD-reduced ESN consistently achieves a higher MC.
Of course, the better performance is justifiable because a POD-reduced ESN is still more structurally complex (originated from a larger ESN) than an ESN (randomly generated) with the same number of neurons as the reduced network.

\subsubsection{DEIM Reduction}

We also performed DEIM for each POD-reduced ESN to further reduce the number of $\tanh$ in the computations and evaluate the drop in MC compared to the POD-reduced ESN.
We tested four different energy cutoff configurations for the DEIM: $\{1\%,5\%,10\%,20\%\}$.
This choice of four values is justified because they represent distinct magnitudes of energy cutoff, testing how the DEIM behaves on four different approximation precision requirements.
    
\begin{table*}[]
    \centering
    \caption{
    Memory capacity evaluated for different energy cutoffs used in POD and DEIM. Each table considers an original ESN with a different size $N$, to be reduced.
    }
    \begin{tabular}{|c|c|c|c|c|c|}\hline
        $N=800$  & \multicolumn{5}{c|}{Energy Cutoff (EC) for DEIM} \\\hline
        EC (POD) & $0\%$ &$1\% (678)$ & $5\% (430)$ & $10\% (279)$ & $20\% (128)$ \\\hline
        $0\% (800)$ & $19.88 \pm 0.01$ &$-$ & $-$ & $-$ & $-$ \\\hline
        $1\% (686)$ & $19.87 \pm 0.01 $ & $0.44 \pm 0.20$ & $0.099 \pm 0.01$ & $0.08 \pm 0.04$ & $0.54 \pm 0.18$ \\\hline
        $5\% (445)$ & $19.86 \pm 0.01 $ &$ 16.48 \pm 2.21$ & $0.059 \pm 0.026$ & $0.096 \pm 0.02$ & $0.55 \pm 0.17$ \\\hline
         $10\% (291)$ & $19.84 \pm 0.008 $ &$19.68 \pm 0.03$ & $0.99 \pm 0.25$ & $0.097 \pm 0.02$ & $0.54 \pm 0.18$ \\\hline 
         \multicolumn{6}{c}{}
    \end{tabular}\\
    \begin{tabular}{|c|c|c|c|c|c|}\hline
        $N=1,400$  & \multicolumn{5}{c|}{Energy Cutoff (EC) for DEIM} \\\hline
        EC (POD) & $0\%$ &$1\% (1,186)$ & $5\% (748)$ & $10\% (484)$ & $20\% (226)$ \\\hline
        $0\% (1,400)$ & $19.93 \pm 0.003$ &$-$ & $-$ & $-$ & $-$ \\\hline
        $1\% (1,119)$ & $19.93 \pm 0.003$ &$0.11 \pm 0.03$ & $0.03 \pm 0.03$ & $0.04 \pm 0.02$ & $0.17 \pm 0.05$ \\\hline
        $5\% (772)$ & $19.91 \pm 0.003$ &$3.189 \pm 1.09$ & $0.03 \pm 0.02$ & $0.04 \pm 0.02$ & $0.17 \pm 0.04$ \\\hline
         $10\% (505)$ & $19.90 \pm 0.003$ &$19.18 \pm 0.50$ & $0.19 \pm 0.02$ & $0.025 \pm 0.02$ & $0.17 \pm 0.05$ \\\hline 
         \multicolumn{6}{c}{}
    \end{tabular}\\
    \begin{tabular}{|c|c|c|c|c|c|}\hline
        $N=2,000$  & \multicolumn{5}{c|}{Energy Cutoff (EC) for DEIM} \\\hline
        EC (POD) & $0\%$ &$1\% (1,835)$ & $5\% (1,122)$ & $10\% (713)$ & $20\% (333)$ \\\hline
        $0\% (2,000)$ & $19.96 \pm 0.002$ &$-$ & $-$ & $-$ & $-$ \\\hline
        $1\% (1,682)$ & $19.95\pm 0.002$ &$0.06 \pm 0.01$ & $0.02 \pm 0.02$ & $0.03 \pm 0.01$ & $0.03 \pm 0.03$ \\\hline
        $5\% (1,045)$ & $19.94\pm 0.002$ &$1.2 \pm 0.4$ & $0.01 \pm 0.02$ & $0.02 \pm 0.02$ & $0.08 \pm 0.03$ \\\hline
         $10\% (671)$ & $19.92\pm 0.002$ &$18.37 \pm 0.90$ & $0.09 \pm 0.03$ & $0.04 \pm 0.01$ & $0.07 \pm 0.03$ \\\hline 
    \end{tabular}\\
    \label{tab:deim_results}
\end{table*}

Table \ref{tab:deim_results} shows the results of applying these DEIM configurations into each POD for three original reservoir sizes $N = \{800,1400,2000\}$ (from the topmost table to the bottom-most one, respectively).
It presents the results for the DEIM reduction, where the memory capacity is evaluated for each configuration in energy cutoff for both POD and DEIM. The number in parenthesis is the actual dimension resulting from the reduction. Each column corresponds to a different energy cutoff configuration for DEIM, evaluated in the first row. In contrast, each row represents a different energy cutoff configuration for POD, evaluated in the first column. 
For instance, the MC of an ESN with a $1\%$ energy cutoff POD (yielding $1,119$ states when $N=1,400$) and a $5\%$ energy cutoff DEIM (yielding $748$ $\tanh$ function evaluations when $N=1,400$) is 0.099, 0.03 and 0.02 for $N=800, 1400, 2000$ respectively. Notice that there was no POD reduction for the first row of each table and no DEIM reduction for the first column of each table. The empty cells indicate that DEIM can not be employed without first applying the POD reduction.
The only time DEIM achieved an MC close to the MOR was when there was a $1\%$ energy cutoff for DEIM considering $10\%$ energy cutoff for POD.
That is, DEIM is performed for smaller reduced-order models.
Regarding the experiments, performance is generally mildly better whenever DEIM has a higher number of states ratio than the POD states.
For this experiment, DEIM did not perform well as expected since the white noise signal does not allow for a significant reduction of states, as it is a highly heavy information signal.

\subsection{NARMA System}

As an initial case study for the POD reduction of the ESN, we try to identify the behavior of a so-called NARMA (Nonlinear Autoregressive Moving Average) difference equation system \cite{Sakemi2020}, equated as follows:
\begin{multline} \label{eqn:narma}
    y[k] = 0.3y[k-1] +0.05y[k-1]\sum_{i=1}^m y[k-i] \\+ 1.5u[k-m+1]u[k] + 0.1
\end{multline}
where $m=10$ is the order of the system.

As in \cite{Sakemi2020}, the excitation signal applied in \eqref{eqn:narma} is drawn from the random uniform distribution with a value range of $0 \leq u[k] \leq 0.05$.
A simulation performs $5,000$ time steps where the first 2,000 samples are labeled as training data and the rest is labeled as test data.
This work employs the $R^2$ metric to measure network performance.

With the dataset mentioned above, we train an ESN with the following configuration: 
$1,400$ neurons in the reservoir layer, high enough so that we show the MOR potential at work;
a leak rate of $\gamma = 0.7$; 
  scaling of $0.1$ for both bias and input connections; and spectral radius of $\rho = 0.99$.
In terms of $R^2$, the network had a performance of $0.95949337$ for the NARMA model output.
We will now carry out experiments of POD for this network to evaluate how the MOR performs in terms of $R^2$ concerning the original $1,400$ units network.
\begin{figure}
    \centering
    \includegraphics[width=0.85\textwidth]{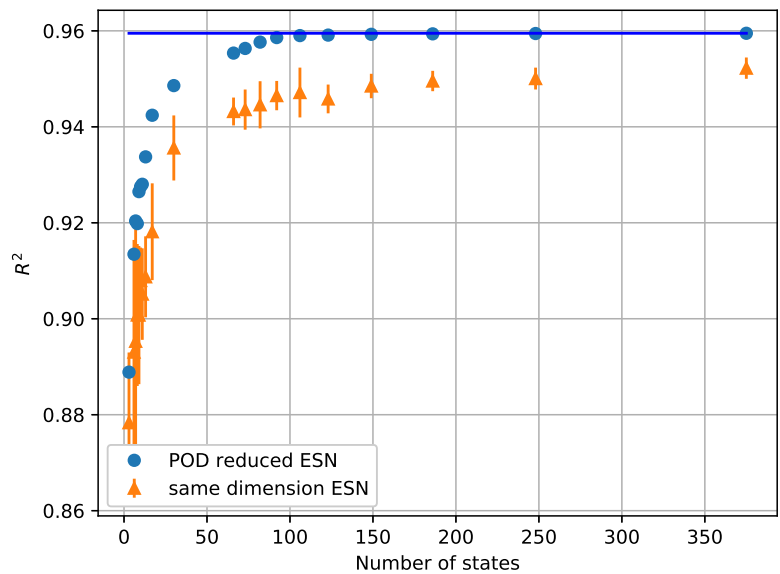}
    \caption{Experiment 
    comparing a POD-reduced ESN (blue dots) with an ESN of equivalent size (to the reduced ESN) (orange triangles) for the 10th-order NARMA task. The POD reduction is applied on an ESN with $1,400$ units in the reservoir.
   The horizontal axis is the number of states (units) of the reduced (full) network, while the vertical axis is the $R^2$ metric on the test set.
    The plot's blue horizontal line corresponds to the $R^2$ of the $1,400$ units ESN.
    }
    \label{fig:pod_narma}
\end{figure}

Figure \ref{fig:pod_narma} showcases the experiment regarding applying POD reduction so that the number of states of the POD-reduced ESN appears in the $x$ axis (blue dots).
For comparison, we also plotted the $R^2$ for the same NARMA experiment with $10$ runs of full (non-reduced) ESNs with the same reservoir size as the networks that underwent POD reduction (orange triangles).
The POD-ESN reduction generally achieved superior performance over the full ESN at the same \eric{reservoir} size, which is understandable, as the POD-reduced ESN is not only supposed to be an emulation of a larger ESN behavior but also more complex in structure.
The NARMA experiment also shows that the $R^2$ metric for ESNs reduced to at least $50$ states is very similar to the metric achieved by the original 1,400 units ESN, i.e., the blue dots are very close to the horizontal blue line in the plot of Figure \ref{fig:pod_narma} when the number of states is higher than 50.

\subsection{Two Wells and One Riser Platform}

\begin{figure}[!b]  %
	\centering
	\includegraphics[width=0.8\textwidth]{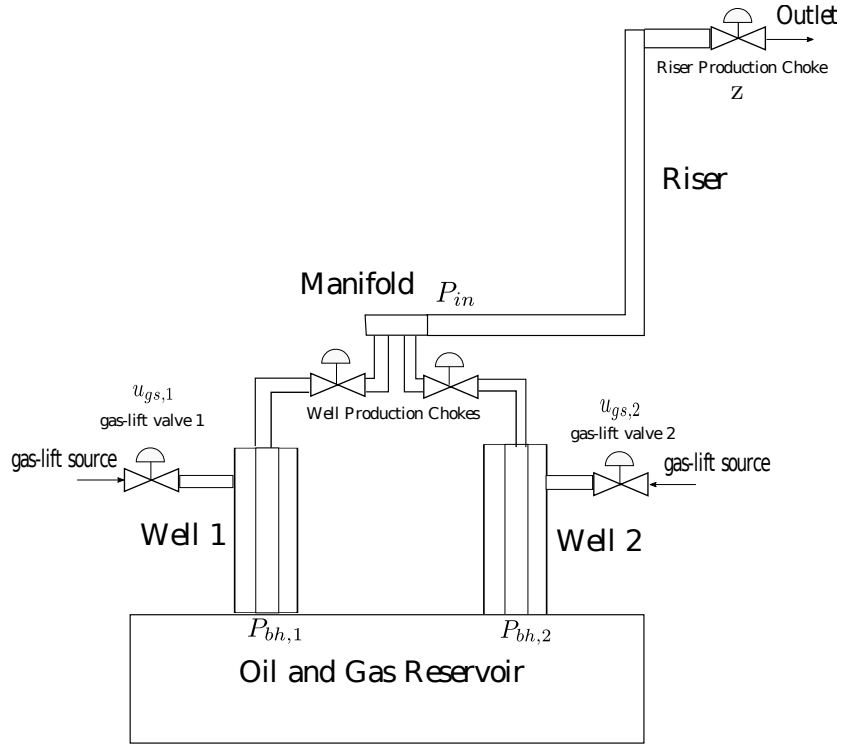}
	\caption{Representation of an oil platform containing two wells and one riser. 
		From \cite{jordanou_eaai}.
	}
	\label{fig:schematic}
\end{figure}

We now test the MOR over the ESN for a physical problem: an oil production platform consisting of two gas-lifted oil wells and one riser, as illustrated in Figure \ref{fig:schematic}.
To gather data, we utilize a composite model consisting of two well models, a riser model, and a manifold that connects the three units.

All models assume a 2-phase fluid containing gas and liquid.
The well model assumes two control volumes in the gas injection annulus and the production tubing, with boundary conditions for gas-lift, reservoir, and outlet pressure.
The riser model considers a horizontal pipeline and the vertical portion of the riser as two separate control volumes while assuming the inlet flow and outlet pressure as boundary conditions.
The manifold assumes no load loss due to friction; therefore, it equates the sum of the output flow from the wells to the riser input flow and the output pressure of each well to the riser inlet pressure.

Overall, the system has 120 algebraic variables, 10 state variables, 5 input variables, and precisely 5 boundary conditions.
\cite{esmaeil} presents the model in more detail, while \cite{jahanshahi2011} describes the riser model .
The model configuration is the same as the one described in \cite{jordanou_ieee}.
The reader can refer to these works for more details on the mathematical modeling of the platform.

The experiment with the two-well production platform depicts how to achieve MOR \eric{with ESNs} from a system identification standpoint.
First, we must train an ESN model for the two-well one-riser platform.
We generate $50,000$ timesteps of data from numerical simulation of the platform model, yielding a dataset where the 2-dimensional input to the ESN is composed of both well-production chokes $u_{ch,1}$ and $u_{ch,2}$. Further, the desired 2-dimensional output of the network corresponds to each well bottom-hole pressure: $P_{bh,1}$, $P_{bh,2}$.
The training dataset consists of the first $10,000$ timesteps, while the segment from $k = 20,000$ to $k = 30,000$ serves as a validation set, and the rest ($k > 30,000$) as a test set.
With the described dataset, we train an ESN with $1,400$ reservoir units (chosen this high for the sake of demonstrating the MOR potential at work), a leak rate of $\gamma = 0.7$, scalings for both bias and input equal to $0.1$, and spectral radius $\rho = 0.99$.
In terms of $R^2$ metric, the network had a test performance of $(0.99881673,0.99900379)$ for each \eric{individual} well bottom-hole pressure.

Now, we run POD experiments with the previously trained network to assess how MOR performs in terms of $R^2$ compared to the original $1,400$ units network.
\begin{figure}
    \centering
    \includegraphics[width=0.85\textwidth]{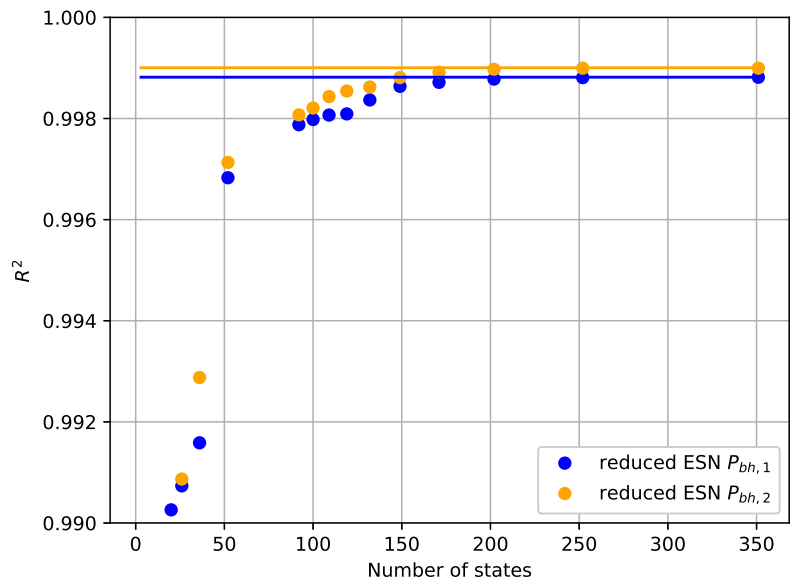}
    \caption{POD-ESN for a system identification task.
    The full ESN network has $1,400$ neurons and was trained to model the platform with two wells and one riser.
    The $x$ axis is the number of states of the reduced network, whereas the $y$ axis is the $R^2$metric on the test set for each output variable (bottom-hole pressures). The bottom-hole pressure of the first well is represented in blue, while the orange color denotes the bottom-hole pressure of the second well.
    The $R^2$ of the original network corresponds to the horizontal lines of the respective colors for comparison.}
    \label{fig:pod_twowells}
\end{figure}
Figure \ref{fig:pod_twowells} depicts an experiment where MOR of different state sizes was tested in terms of $R^2$ over the test data.
One can infer that, after a given number of states $(150)$, the performance remains consistently close to the original network in terms of $R^2$, despite having only $10\%$ of the original number of states.

POD reduction that resulted in $92$ states also showcased good performance compared to the original network of $1,400$ neurons.
However, with only POD, the computational problem of computing $\mathbf{T}^T\mathbf{f}$ remains.
We select the case where the reduced network has $92$ states (representing an energy cutoff of  $1\%$) and try performing DEIM on it.
\begin{figure}
    \centering
    \includegraphics[width=0.85\textwidth]{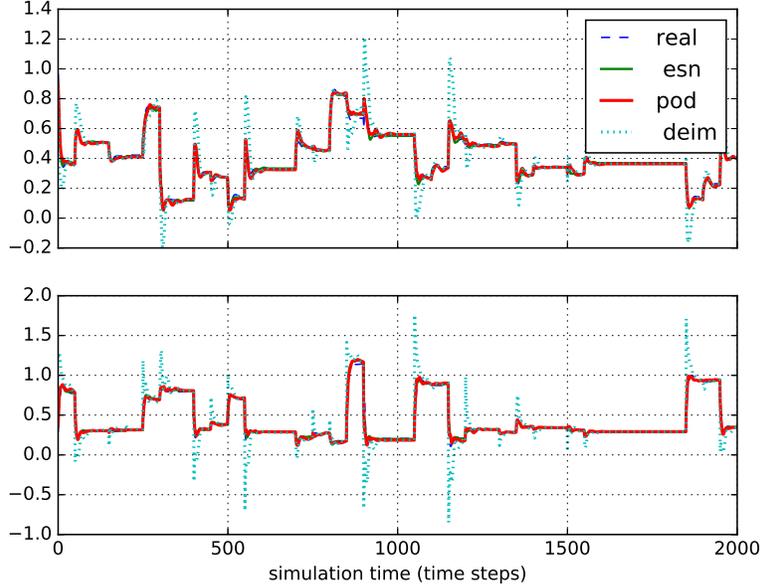}
    \caption{Single simulation run involving a POD with $92$ states ($0.01$ energy cutoff) and a DEIM interpolation with $m=1,073$, put side by side with the original data for the bottom hole pressure $p_{bh}$ of both wells (normalized), and the original ESN.}
    \label{fig:twowells_plot}
\end{figure}
Figure \ref{fig:twowells_plot} depicts a simulation for the ESN, POD-ESN, and POD-DEIM ESN for
the test data of the two-wells and one riser platform.
Even though there was a reduction from $1,400$ to only $92$ states, the behavior of the ESN and the POD-ESN managed to be close in terms of dynamics.
The application of DEIM reduced the computation nodes from $1,400$ to $1,073$; however, some overshooting emerged, which was not present in the ESN and POD-ESN.
Concerning the simulation run in Figure \ref{fig:twowells_plot}, the $R^2$ for the normalized bottom-hole pressure of each well was:
$(0.9988,0.9990)$ for the ESN, 
$(0.9979,0.9981)$ for the POD-ESN, 
and $(0.9873,0.9671)$ for the DEIM-POD-ESN.
%
There is little drop in response quality from reducing the number of states from $1,400$ to $92$ through POD, but performing interpolation from a standard POD to a POD-DEIM framework seems to affect the response more significantly.
The small drop in response quality concerning the POD-ESN is expected, as the POD was performed requesting a 1\% energy cutoff.
In other words, the reduced-order model is $99\%$ close to the original ESN regarding dynamic information.



\section{Discussion}

POD-reduced ESN achieved a response close to the original ESN for the NARMA and the two-well one-riser case study, while it incurred a minor performance loss in the MC experiments. However, DEIM did not reach the same performance as POD in those experiments.  
These findings indicate that DEIM incurs more dynamic-information loss than POD, as the latter retains the number of activation functions in the reduced model even though it reduces the number of states.
Thus, we conjecture that the capacity of a reservoir to represent a nonlinear system accurately is more influenced by the combination of the nonlinear functions in a high-dimensional space than by maintaining a high-dimensionality of the reservoir states themselves. In the context of MOR, this function combination is given by lifting the reduced states back to the original space just before applying the $\tanh$ nonlinearity.

The application of POD leads to some reduction in the memory required for storing and using the POD-reduced ESN.
First,  the state-to-output linear combination matrix $\mathbf{W_r^oT}$ maps the reduced space directly to the output, invariably reducing its size.
Also, the computation of the activation functions becomes slightly less expensive memory-wise because the resulting matrix $\mathbf{W_r^rT}$, which is a product computed offline, has fewer elements.
Of course, the resulting matrix is still large compared to an ESN with the same size as the reduction, rendering the same-size ESN less complex than the POD-reduced one.

Even though POD computes the same number of activation functions as the original ESN, the computation time is significantly reduced, as shown in Table \ref{tab:narma_time}.
This table shows the mean time it took to execute a step in the full ESN against the time it took to perform a POD-ESN computation step for the NARMA experiment.
For instance, when applying POD-ESN to reduce from 1400 states to 66 states, we get an 80\% decrease in mean execution time (from 0.767 ms to 0.147 ms) while still maintaining excellent performance, as this setup is near the horizontal line in Figure ~\ref{fig:pod_narma}.
All experiments were performed under similar conditions and with the same computer.

\begin{table}[h!]
    \centering
    \caption{Mean execution time for the NARMA experiment composed of $5,000$ time steps.}
    \begin{tabular}{|c|c|c|}\hline
         & Mean Execution Time (ms)& St. Dev. (ms)\\\hline
         ESN\;(size=1400) & 0.767 & 0.537\\\hline
         POD\;(size=3) & 0.072 & 0.0498\\\hline
         POD\;(size=6) & 0.078 & 0.0251\\\hline
         POD\;(size=7) & 0.141 & 0.391\\\hline
         POD\;(size=8) & 0.131 & 0.340\\\hline
         POD\;(size=9) & 0.160 & 0.543\\\hline
         POD\;(size=10) & 0.140& 0.467\\\hline
         POD\;(size=11) & 0.233 & 0.955\\\hline
         POD\;(size=13) & 0.105 & 0.122\\\hline
         POD\;(size=17) & 0.106 & 0.0738\\\hline
         POD\;(size=30) & 0.135 & 0.0856\\\hline
         POD\;(size=66) & 0.147 & 0.254\\\hline
         POD\;(size=73) & 0.144 & 0.138\\\hline
         POD\;(size=82) & 0.141 & 0.140\\\hline
         POD\;(size=92) & 0.155 & 0.109\\\hline
         POD\;(size=106) & 0.151 & 0.0744\\\hline
         POD\;(size=123) & 0.149 & 0.0907\\\hline
         POD\;(size=149) & 0.183 & 0.407\\\hline
         POD\;(size=186) & 0.201 & 0.145\\\hline
         POD\;(size=248) & 0.230 & 0.134\\\hline
         POD\;(size=375) & 0.408 & 0.199\\ \hline
    \end{tabular}
    \label{tab:narma_time}
\end{table}

As shown in Table \ref{tab:narma_time}, even though there is no computation reduction in the nonlinear nodes, the computational time for a POD-ESN to compute a time step is reduced, even if by a small margin.
This computation-speed gain happens precisely because the reduced-order ESN has fewer states, despite the nonlinear node computation remaining unchanged.

As previously discussed, the poor performance of DEIM in the memory capacity experiments corroborates the loss of stability incurred in the DEIM-reduced ESNs.
Besides, even when the DEIM-reduced ESN dynamic system remained stable, as in the two-well experiment illustrated in Figure \ref{fig:twowells_plot}, the system experienced high overshoots translating into modeling errors.
The independent work \cite{Wang2022} that also implements POD/DEIM on ESN, which appeared in the literature during the writing of this research, proposes a method to deal with the stability issue.
However, the method is restricted to the particular class of ESNs with dynamic equations without the bias term.
That method relies on expanding the nonlinear dynamics reduced by the DEIM so that the Jacobian contribution of the terms affected by $(\mathbf{P}^T\mathbf{U})$ becomes null concerning $\mathbf{u}=\mathbf{0}$.
In this context, generalized methods (which account for the bias term as well) to guarantee stability retention of an ESN interpolated by DEIM are an interesting topic for future works.

\section{Conclusion} \label{sec:conclusion}

In this investigation, the POD achieved exceptional results in reducing the number of states of an ESN and maintaining performance.
The reduced ESN performed nearly as well as the original ESN, despite the drastic reduction of states in a typical system identification task.
This work also showcased how the nature of the excitation signal changes the singular value profile of the SVD, concluding that lower-frequency input signals can result in more efficient reductions. Ideally, the excitation signal should be as slow as necessary to identify a system.

However, despite performing MC tests considering signals that carry information from all frequencies, the POD-reduced network performed better than an ESN of the same size trained on the data.
Arguably, the superior performance of the POD-reduced ESN may be attributed to its ability to emulate the behavior of the larger original ESN. Additionally, the increased complexity of the reduced network, compared to an ESN of the same size, could contribute to its enhanced performance.

These findings imply that applying POD to reduce the number of states (reservoir size) of an ESN is an excellent strategy to obtain a smaller model that behaves almost equivalently to the original one. However, some adaptation to the DEIM method may be necessary before it can be applied to increase model efficiency further.
Also, reducing the reservoir size using POD has the advantage of interpretability since the states are sorted and pruned according to the energy contribution metric.
Finally, applying POD to an ESN can show which linear combination of states contributes more significantly to the ESN dynamic behavior.

For possible future work, we will test the developed POD-ESN model in predictive control applications, comparing the performance of the reduced-order model to its full-order counterpart.
Further, there are applications in reservoir computing, such as time series prediction problems, which could benefit from a reservoir reduction using the POD-ESN.
 Another direction for future research is the study of ways to adapt DEIM to perform model reductions more consistently.

\section*{Acknowledgments}
This work was funded in part by FAPESC (grant 2021TR2265), CAPES  (grant 88882.182533/2011-01), and CNPq (grant 308624/2021-1).



\bibliographystyle{elsarticle-num-names} 
\bibliography{references}


\end{document}